\documentclass[runningheads]{llncs}
\usepackage[T1]{fontenc}
\usepackage{graphicx}
\usepackage{amsmath}
\usepackage{amssymb}
\usepackage{algorithm}
\usepackage{algpseudocode}
\usepackage[acronym]{glossaries}
\usepackage{booktabs}
\usepackage{subcaption}
\usepackage{makecell}
\usepackage[figuresright]{rotating}
\newacronym{fl}{FL}{Federated Learning}
\newacronym{ml}{ML}{Machine Learning}
\begin{document}
\title{Federated Learning for Distributed CNC Tool Wear Prediction}
%
%
\author{
Afsana Khan\thanks{Corresponding author} \and
Morris Stallmann \and
Marcin Pietrasik \and
Charalampos~S.~Kouzinopoulos \and
Anna Wilbik
}

\authorrunning{A. Khan et al.}

\institute{Maastricht University, Maastricht, Netherlands \\
\email{a.khan@maastrichtuniversity.nl}}

\maketitle              
\begin{abstract}
Tool wear prediction is an important task in CNC machining, where accurate monitoring of tool condition supports product quality and process reliability. Machine learning methods have shown potential for this task, but their use in industrial environments is limited by the distributed nature of machining data and by restrictions on data sharing between machines, sites, or organizations. Federated learning offers a suitable framework for this setting by enabling collaborative model training without transferring raw operational data. However, it is open if federated learning can lead to accuracy gains in CNC tool wear prediction that justify the increased complexity of such a system. In this experimental study, real tool trajectories are distributed across simulated clients to represent a federated learning scenario. The federated models are compared against centralized references and local client baselines. Results show that federated learning achieves performance close to centralized learning and improves significantly over local client models. These findings indicate that federated learning can support collaborative tool wear prediction in distributed CNC manufacturing environments and the increased complexity is justified.

\end{abstract}

\section*{Introduction}

Cutting tools are essential in machining processes such as milling, drilling, and sawing, but are affected by wear due to abrasion, adhesion, and diffusion, which can eventually lead to tool failure or defective parts \cite{MATIVENGA2024212,Patel2026-gg}. Machine learning (\acrshort{ml}) methods have shown promising performance for tool-wear prediction \cite{soori_cnc_machine_tools,shah_ml_wear_prediction}, but reliable models require sufficiently diverse data. In realistic CNC environments, such data are distributed across machines, production cells, factories, or organizations, while centralization may be costly or undesirable because sensor recordings and images are large and may reveal production conditions, machining strategies, material properties, or process quality. \acrfull{fl} provides an alternative by enabling collaborative training while keeping raw operational data local \cite{kairouz_federated}.
Previous studies have investigated \acrshort{fl} for tool-wear prediction. Kaleli et al. \cite{kaleli_domain_aware} combine feature extraction with a federated (Bi)LSTM model on the PHM dataset \cite{phm_dataset}, but do not establish whether federated training improves over isolated local training. Huang and Lee \cite{HUANG2024150} address client heterogeneity through a new aggregation method using the dataset of \cite{lin_tool_wear}, but focus on aggregation rather than distributed preprocessing; the original study applies normalization to raw data \cite{lin_tool_wear}, which may be difficult when data cannot be centrally shared. De Melo Rosa et al. \cite{DEMELOROSA2026415} demonstrate image-based federated tool-wear prediction and improvements over per-client training, but acknowledge the limited generality of their small-scale evaluation. Thus, it remains unclear whether \acrshort{fl} can approach centralized performance while improving over local-only learning under distributed data-processing constraints. We investigate this question using the Multimodal Automatic Tool Wear Inspection (MATWI) dataset \cite{de2023matwi}. We construct federated sensor- and image-based prediction pipelines in which complete tool trajectories remain local to clients and compare federated training against centralized and local-only baselines.

\section*{MATWI Dataset and Preprocessing}\label{sec:dataset}

The MATWI dataset contains 17 cutting-tool sets, each representing the wear progression of one milling insert. Each machining cycle includes wear labels, synchronized force, accelerometer and acoustic recordings, and an image of the cutting edge. Wear is categorized as flank wear, adhesive wear, or combined flank-and-adhesive wear. Following \cite{de2023matwi}, sets 1-13 are used; sets 14-17 are excluded due to a workpiece-material domain shift. Tool sets remain intact when assigned to clients.
Sensor recordings contain air-cut intervals, where the tool moves without engaging the workpiece, and active-cutting intervals; only the latter are retained. A level-four discrete wavelet decomposition using the Daubechies 4 wavelet \cite{mallat1989theory,daubechies1990wavelet} is applied independently to two sensor channels:
\[
E[k]=|D_1[k]|, \qquad
z[k]=\frac{E[k]-\mu_E}{\sigma_E+10^{-9}},
\]
where $D_1[k]$ is the level-one detail coefficient at index $k$, $E[k]$ its magnitude, $z[k]$ the normalized activity score, and $\mu_E,\sigma_E$ the mean and standard deviation of the first 1000 $E[k]$ values. Samples with $z[k]>10$ are considered active and the detected interval is expanded by 5\%. For channel intervals $[s_1,e_1]$ and $[s_2,e_2]$, the retained interval is $[\max(s_1,s_2),\min(e_1,e_2)]$. If fewer than 1000 active samples are detected for a channel, the complete recording is retained.

\begin{figure}[h]
    \centering

    \begin{subfigure}{0.72\linewidth}
        \centering
        \includegraphics[width=\linewidth]{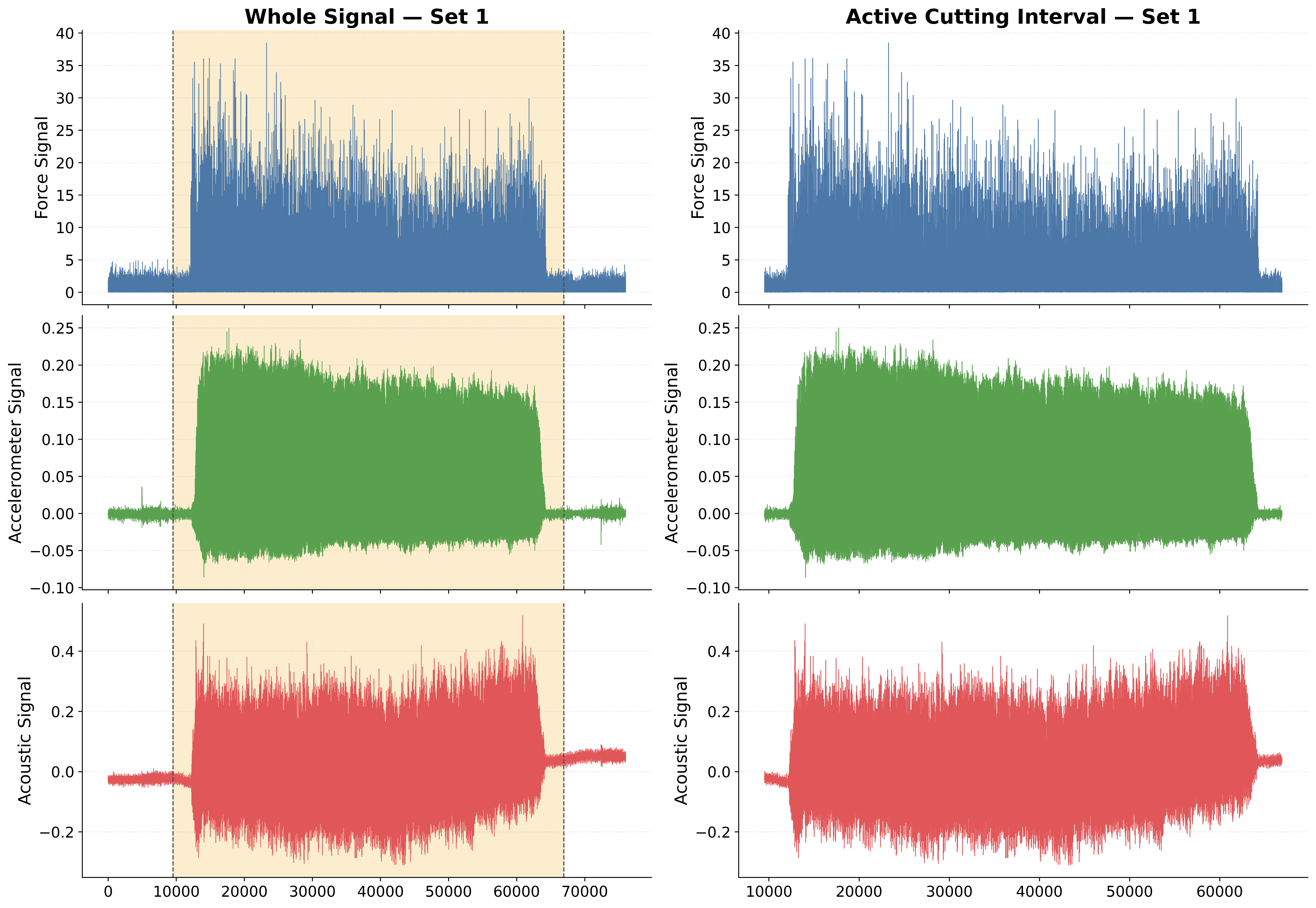}
        \caption{Example sensor recordings before and after air-cut removal.}
        \label{fig:aircut-removal}
    \end{subfigure}

    \vspace{0.3em}

    \begin{subfigure}{0.48\linewidth}
        \centering
        \includegraphics[width=\linewidth]{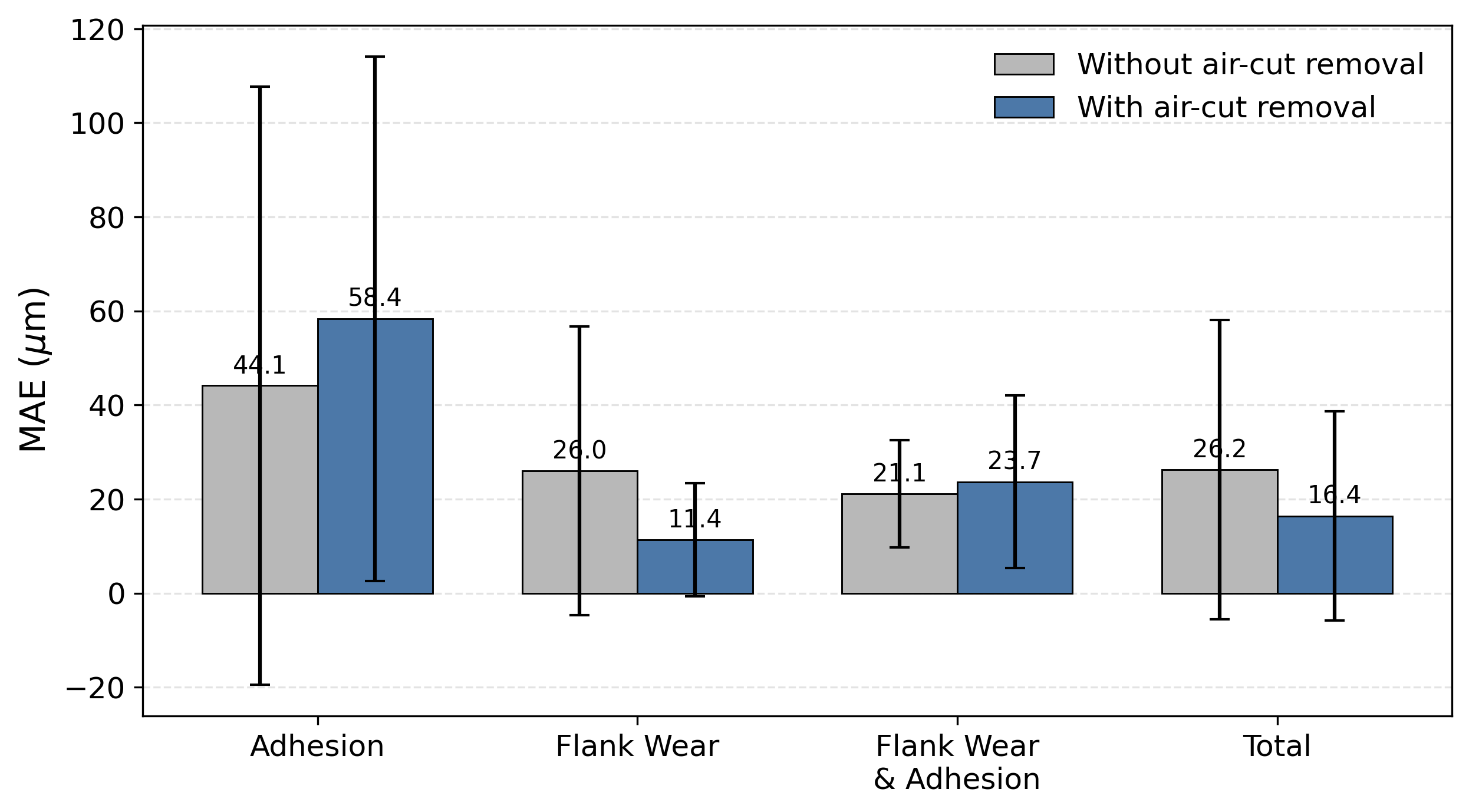}
        \caption{Effect of air-cut removal on model performance with sensor modality.}
        \label{fig:aircut-ablation}
    \end{subfigure}

    \caption{Air-cut removal and its effect on sensor-model performance.}
    \label{fig:aircut}
\end{figure}

Air-cut removal (Figure~\ref{fig:aircut-removal}) reduces the total sensor mean absolute error (MAE) from $26.24~\mu$m to $16.40~\mu$m and flank wear MAE from $26.02~\mu$m to $11.36~\mu$m (Figure~\ref{fig:aircut-ablation}). Adhesion as well as combined flank wear \& adhesion show smaller improvements, likely because they contain fewer samples and adhesion is less consistently reflected in process signals.

\section*{Federated Tool Wear Prediction}\label{sec:method}
We formulate tool wear prediction as a horizontal federated learning problem in which industrial data are generated and retained by separate data owners. Each client represents an autonomous industrial site that holds different samples but shares the same feature space and prediction task. This matches the considered tool wear setting, where different tools or sites contribute data for the same prediction model while keeping their raw sensor recordings or images local.

Training follows a synchronous client-server protocol. At communication round \(t\), the server broadcasts the current global model parameters \(\theta_t\) to all participating clients. Each client initializes its local model with \(\theta_t\), performs local training on its dataset, and returns the updated parameters \(\theta_t^k\) to the server. The server aggregates the client models using FedAvg \cite{mehta2024securing}:

\[
\theta_{t+1}
=
\sum_{k=1}^{K}
\frac{n_k}{n}
\theta_t^k,
\]

where \(K\) is the number of clients, \(n_k\) is the number of samples held by client \(k\), and \(n=\sum_{k=1}^{K}n_k\). Thus, clients with more training samples have proportionally greater influence on the global model.

The same federated training protocol is applied to both sensor and image data. The model architecture and local preprocessing differ by modality, while client coordination, local training, communication, and server-side aggregation remain unchanged.

\subsection*{Experimental Setup}\label{sec:experiments}

Following the MATWI protocol \cite{de2023matwi}, Sets 1, 2, 5, 7, 8, 10, and 11 are used for training, Sets 3, 6, and 12 for validation, and Sets 4, 9, and 13 for testing. The split is performed at tool set level so that cuts from the same physical tool do not occur in multiple splits. The task is regression of the tool wear value for each machining cycle, evaluated under centralized, local, and federated training.

For the sensor modality pipeline, each active cutting signal is divided into overlapping windows of 2048 samples with a stride of 1024, following \cite{yang2022local}. Each window inherits the cut-level wear label, and window predictions are averaged to obtain the cut-level prediction. The 1D-CNN additionally receives a four-dimensional context vector containing the previous wear value and a one-hot encoding of the three wear types. The image modality pipeline following the MATWI dataset paper baseline uses ResNet50. Images are cropped using the set-specific coordinates, resized to $224\times224$ using Lanczos resampling, and normalized using ImageNet mean $[0.485,0.456,0.406]$ and standard deviation $[0.229,0.224,0.225]$.

All models use Adam optimizer, Huber loss with $\delta=1.0$, early stopping with patience 10, and 5 local epochs per federated round. The sensor model uses a learning rate of $10^{-4}$, batch size 32, and 100 federated rounds, while the image model uses $3\times10^{-4}$, batch size 16, and 30 rounds.

The training sets are distributed across three clients: Client A is assigned  Sets 1, 5, and 7; Client B Sets 2 and 10; and Client C Sets 8 and 11. Local models are trained only on their assigned data, while federated training uses the same partition with all three clients participating in every round. Federated training is implemented in Flower 1.30.0 using FedAvg. Performance is reported using mean absolute error (MAE) and the standard deviation of absolute errors for each wear category. 
\section*{Results and Discussion}
\label{sec:results}

\begin{figure}[!ht]
    \centering

    \begin{subfigure}{0.42\linewidth}
        \centering
        \includegraphics[width=0.9\linewidth]{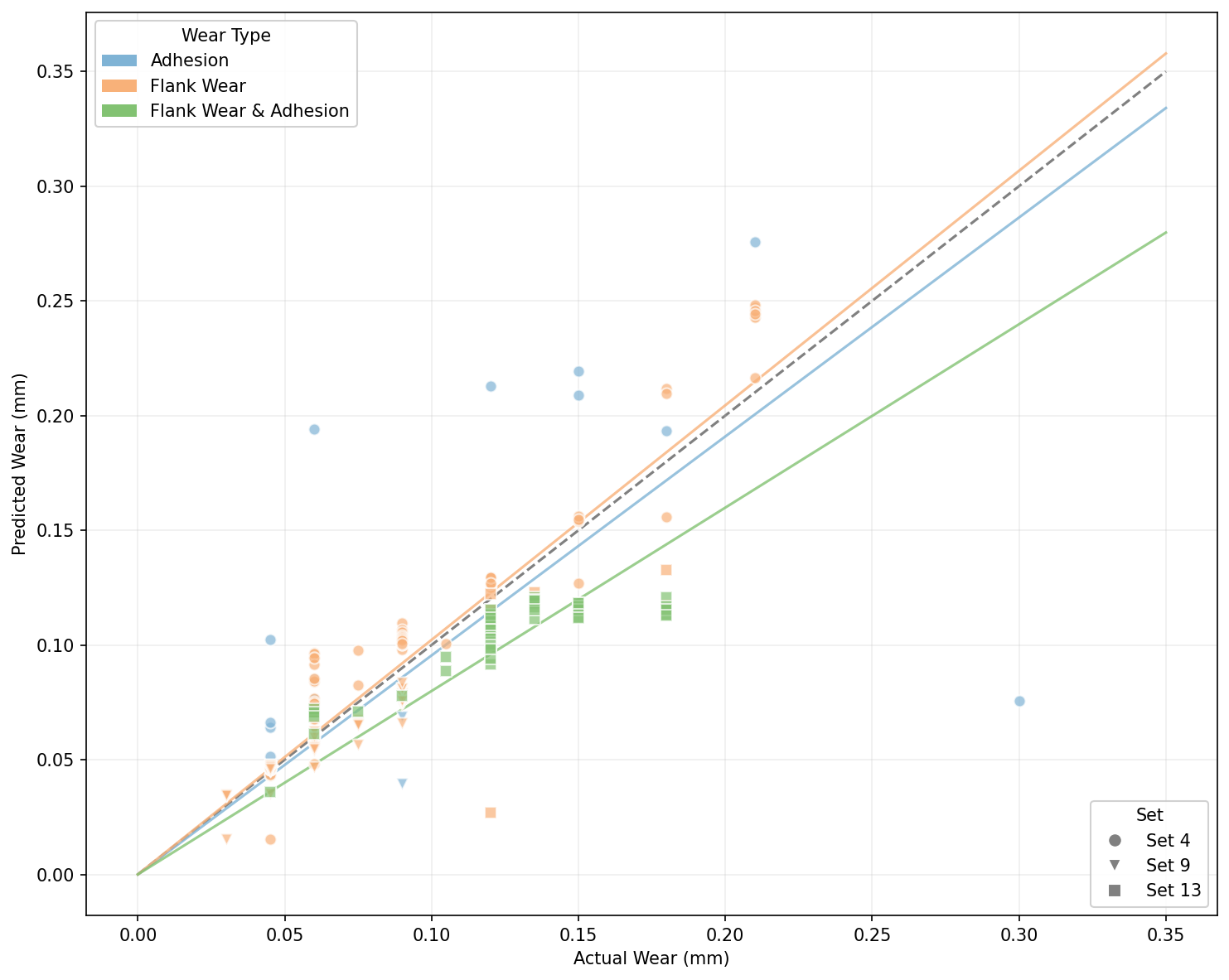}
        \caption{Sensor: Centralized}
        \label{fig:sensor-centralized}
    \end{subfigure}
    \hfill
    \begin{subfigure}{0.42\linewidth}
        \centering
        \includegraphics[width=0.9\linewidth]{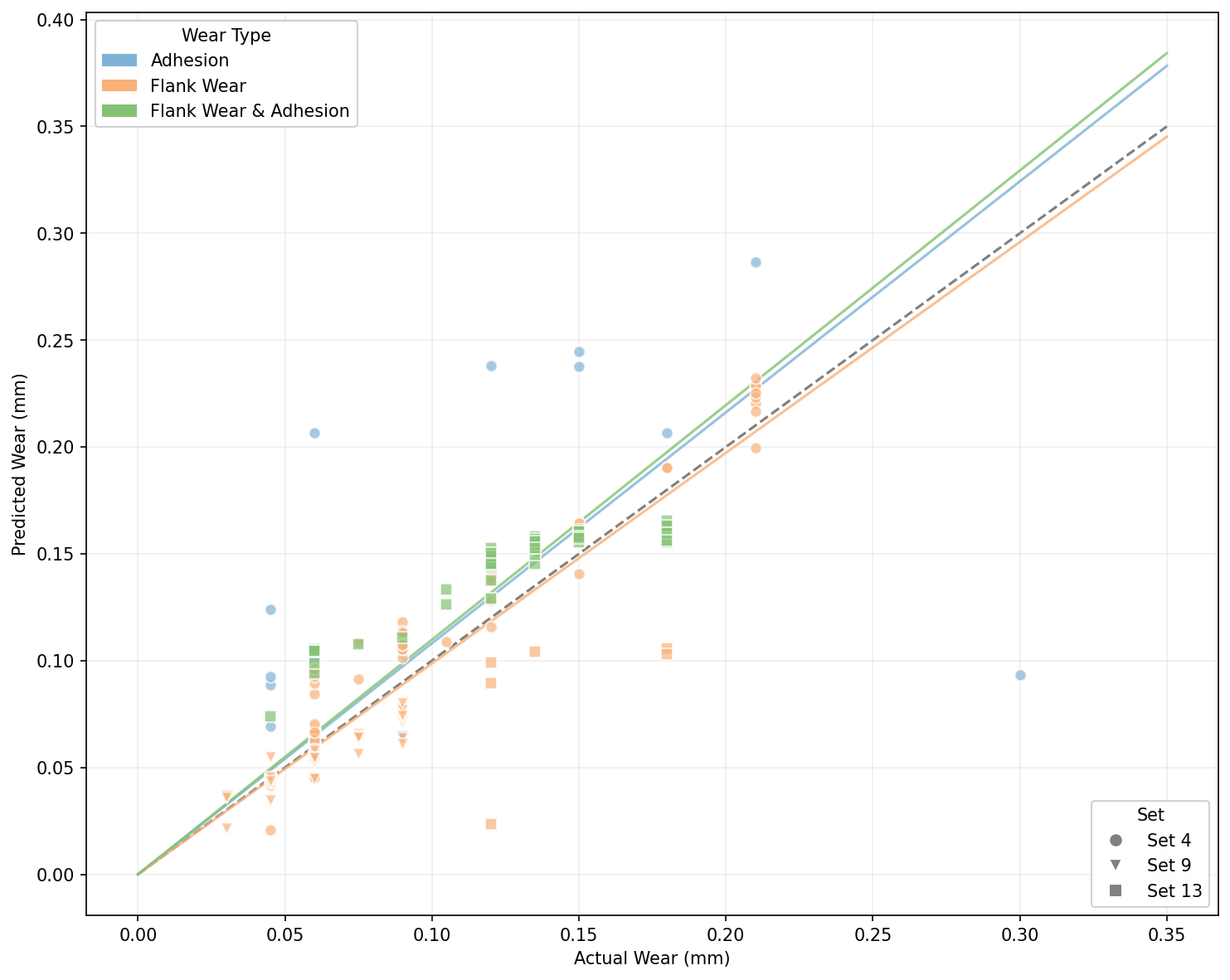}
        \caption{Sensor: Federated}
        \label{fig:sensor-federated}
    \end{subfigure}

    \vspace{0.3em}

    \begin{subfigure}{0.42\linewidth}
        \centering
        \includegraphics[width=0.9\linewidth]{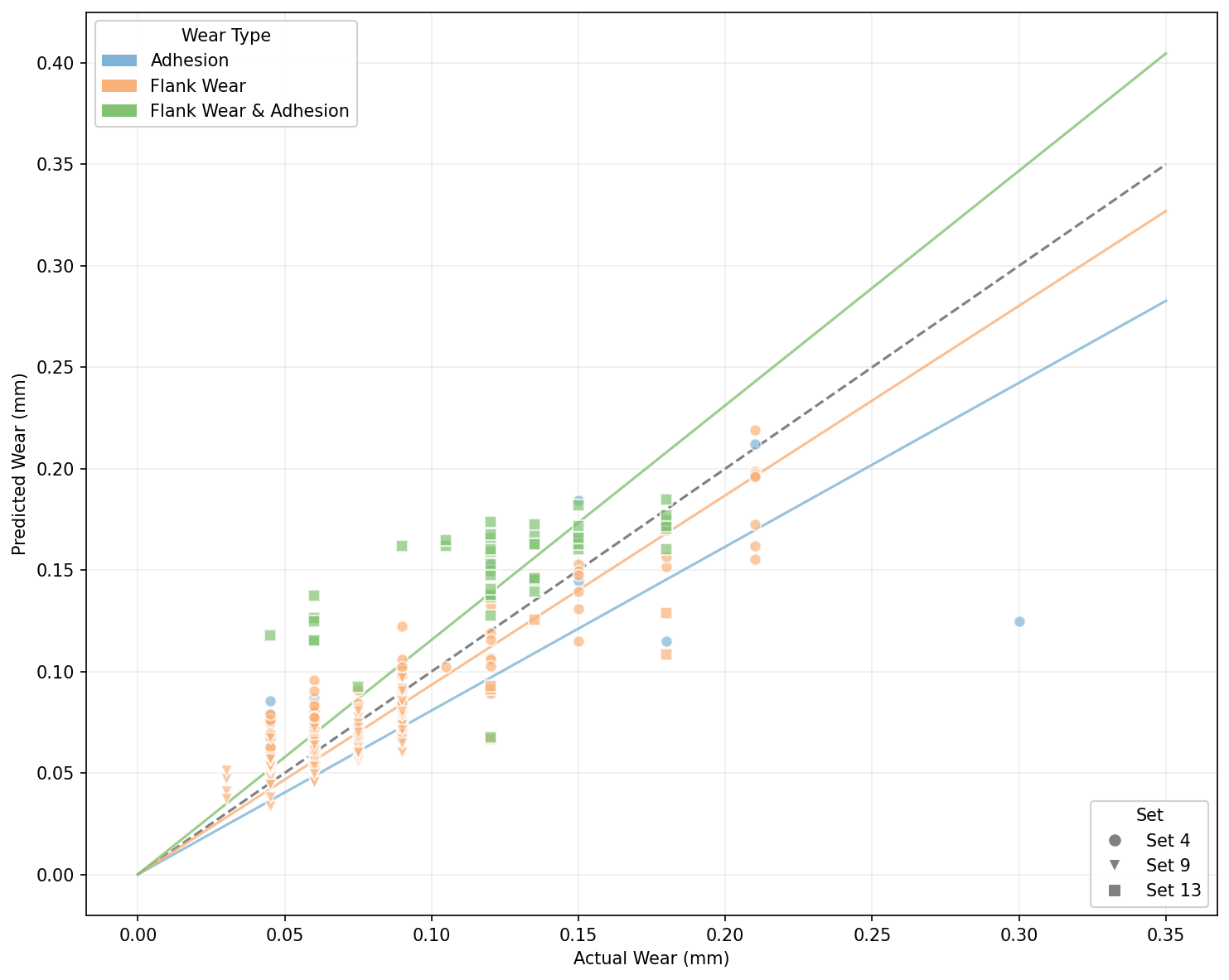}
        \caption{Image: Centralized}
        \label{fig:image-centralized}
    \end{subfigure}
    \hfill
    \begin{subfigure}{0.42\linewidth}
        \centering
        \includegraphics[width=0.9\linewidth]{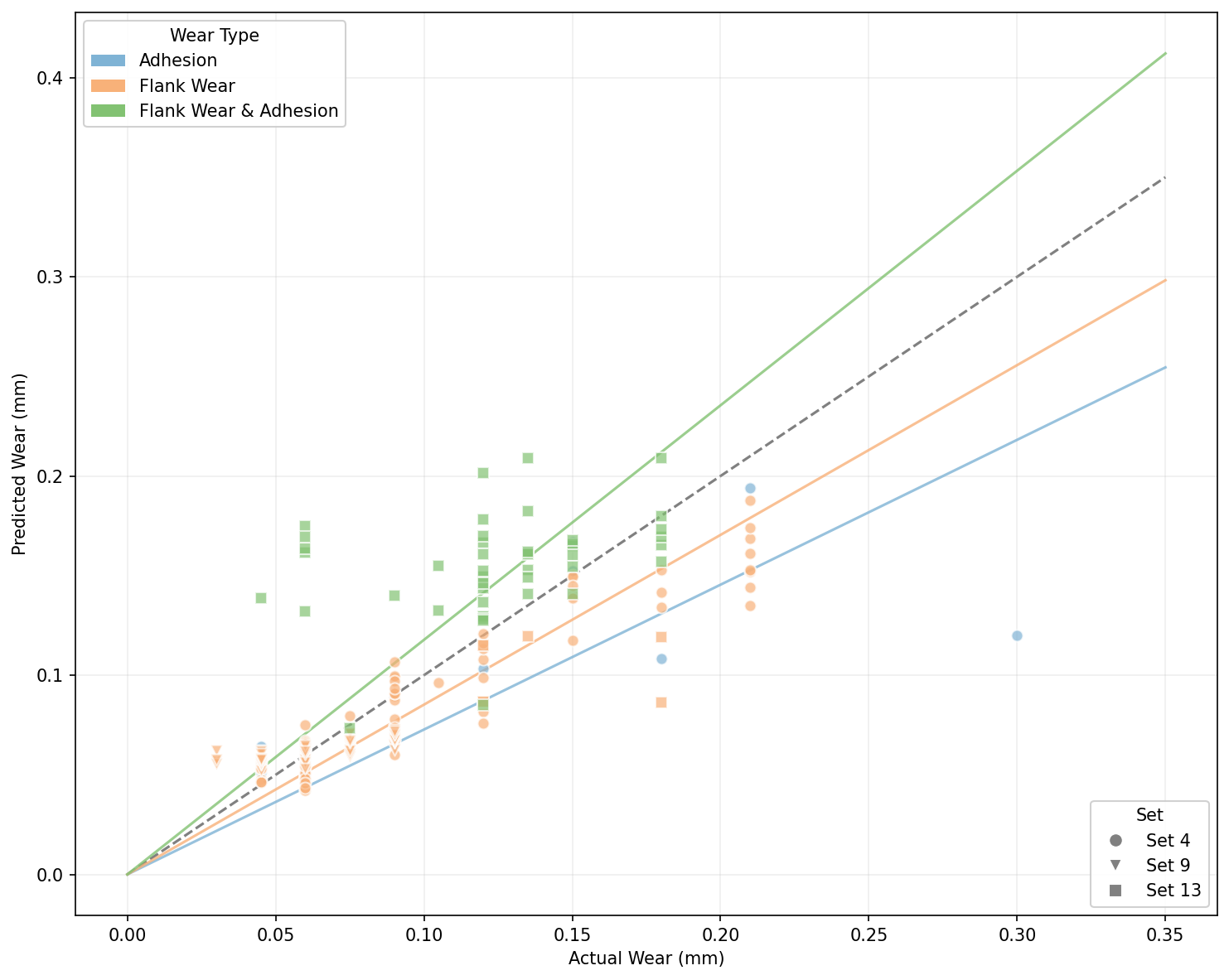}
        \caption{Image: Federated}
        \label{fig:image-federated}
    \end{subfigure}

\caption{Centralized and federated model performance for sensor and image modalities. The dashed gray line denotes ideal prediction ($\hat{y}=y$).}
    \label{fig:main_results}
\end{figure}

Results on test Sets 4, 9, and 13 are reported as MAE in micrometers. Figure~\ref{fig:main_results} shows comparable performance for centralized and federated models within each modality. Flank wear follows the ideal trend more closely (dashed grey line), whereas adhesion shows greater scatter due to fewer samples and weaker representation in process signals whereas in images, adhesion is directly observable. Table~\ref{tab:combined_results} reports MAE and standard deviation for all models.
\begin{table}[!ht]
\centering

\caption{Tool wear prediction performance as MAE $\pm$ standard deviation ($\mu$m). Total denotes MAE over all test samples across wear categories; the best result for each wear type and modality is shown in bold.}
\label{tab:combined_results}
\scriptsize
\setlength{\tabcolsep}{2.5pt}
\renewcommand{\arraystretch}{0.93}

\resizebox{0.9\linewidth}{!}{
\begin{tabular}{llccccc}
\toprule
\textbf{Modality} &
\textbf{Wear Type} &
\textbf{Centralized} &
\textbf{Federated} &
\textbf{Client A} &
\textbf{Client B} &
\textbf{Client C} \\
\midrule

Sensor &
Adhesion &
$58.36 \pm 55.74$ &
$70.75 \pm 52.05$ &
$\mathbf{56.26 \pm 63.40}$ &
$57.62 \pm 40.14$ &
$62.33 \pm 55.78$ \\

&
Flank Wear &
$\mathbf{11.36 \pm 12.04}$ &
$12.50 \pm 12.58$ &
$13.67 \pm 13.39$ &
$27.56 \pm 27.84$ &
$16.63 \pm 20.84$ \\

&
Flank Wear + Adhesion &
$23.66 \pm 18.37$ &
$\mathbf{22.96 \pm 9.70}$ &
$30.27 \pm 19.03$ &
$49.45 \pm 25.64$ &
$62.52 \pm 26.32$ \\

&
\textbf{Total} &
$\mathbf{16.40 \pm 22.24}$ &
$17.90 \pm 22.32$ &
$19.21 \pm 23.95$ &
$33.28 \pm 30.21$ &
$27.58 \pm 32.02$ \\

\midrule

Image &
Adhesion &
$35.19 \pm 40.43$ &
$\mathbf{26.73 \pm 44.21}$ &
$39.17 \pm 52.70$ &
$36.39 \pm 40.06$ &
$39.60 \pm 36.82$ \\

&
Flank Wear &
$\mathbf{13.64 \pm 11.20}$ &
$14.05 \pm 13.98$ &
$28.15 \pm 21.73$ &
$31.43 \pm 23.79$ &
$40.65 \pm 20.41$ \\

&
Flank Wear \& Adhesion &
$\mathbf{31.08 \pm 21.42}$ &
$36.00 \pm 31.38$ &
$34.58 \pm 31.82$ &
$53.61 \pm 30.31$ &
$108.62 \pm 169.62$ \\

&
\textbf{Total} &
$\mathbf{18.06 \pm 18.42}$ &
$18.73 \pm 22.73$ &
$29.97 \pm 26.83$ &
$35.68 \pm 27.62$ &
$52.69 \pm 78.76$ \\

\bottomrule
\end{tabular}
}
\end{table}

For the sensor modality, federated training remains close to centralized training, with total MAEs of $17.90~\mu$m and $16.40~\mu$m, respectively, while outperforming all three local models. Performance is similar for flank wear, while combined wear is slightly better under federated training. For images, federated training achieves $18.73~\mu$m compared with $18.06~\mu$m centrally and also outperforms all local models. It performs similarly on flank wear and better on adhesion.

Adhesion is harder to infer from process signals because it may not produce a clear or consistent signal pattern, whereas in images it is directly visible on the cutting edge. Overall, federated learning approaches centralized performance while improving over isolated training. Remaining differences likely reflect client heterogeneity and the greater variability of the smaller adhesion and combined-wear classes, while the two modalities provide complementary information across wear types.

\newpage
\bibliographystyle{splncs04}
\bibliography{references}

\end{document}